\documentclass{article}

\usepackage[eandd, final]{neurips_2026}

\usepackage[utf8]{inputenc}
\usepackage[T1]{fontenc}
\usepackage{hyperref}
\usepackage{url}
\usepackage{booktabs}
\usepackage{multirow}
\usepackage{amsmath}
\usepackage{amsfonts}
\usepackage{amssymb}
\usepackage{graphicx}
\usepackage{nicefrac}
\usepackage{microtype}
\usepackage{xcolor}
\usepackage{subcaption}
\usepackage{caption}
\usepackage{etoc}
\title{CADMA: Capacity‑Aware Coverage Decomposition for Generative Model Assessment}

\author{
Debin Meng$^{1}$\thanks{Corresponding author: \texttt{debin.meng@qmul.ac.uk}.}\quad
Zheng Gao$^{1}$\quad
Ziquan Liu$^{1}$\quad
Yanran Li$^{2}$\\
\textbf{Ioannis Patras}$^{1}$\quad
\textbf{Georgios Tzimiropoulos}$^{1}$\\[6pt]
{\small $^{1}$Queen Mary University of London}\quad
{\small $^{2}$University of Bedfordshire}\\
}

\begin{document}
\etocdepthtag.toc{main}

\maketitle

\begin{abstract}
Evaluating generative models requires measuring not only sample fidelity but also how well a generated set covers the real data distribution. Recent work has therefore introduced recall-oriented metrics for distribution-level evaluation. However, existing recall metrics often reduce distributional coverage to geometry-based coverage. This simplifying assumption weakens their diagnostic value and interpretability in two ways: these metrics provide only a single aggregate score that cannot distinguish essential diversity failures such as mode dropping and mode collapse, and geometry-based coverage can provide unreliable estimates by assigning high scores to poorly covered generative distributions. We therefore introduce CADMA, a capacity-aware framework for recall-oriented evaluation and diagnosis. CADMA evaluates distributional coverage through local capacity matching between real and generated samples, separating geometry-based coverage from effective coverage. This yields multi-dimensional coverage diagnostics that quantify mode dropping and mode collapse while reducing the overly optimistic estimates produced by geometry-based metrics. Through extensive experiments on toy data, real-world datasets, and generative models, we show that CADMA provides more reliable coverage estimates and more fine-grained, interpretable diagnostic signals of diversity failures than existing metrics. Our code is available at \url{https://github.com/Open-Debin/CADMA_NeurIPS2026}.
\end{abstract}

\section{Introduction}
\label{sec:intro}

Generative models have improved rapidly in recent years, with modern VAE-based models~\cite{kingma2013auto,van2017neural}, GANs~\cite{goodfellowGenerativeAdversarialNets,karrasStyleBasedGeneratorArchitecture2019}, and diffusion models~\cite{rombachHighResolutionImageSynthesis2022,hoDenoisingDiffusionProbabilistic2020,songDenoisingDiffusionImplicit2022} producing increasingly realistic outputs. As these models become stronger, reliable evaluation becomes more important. Unlike discriminative models, which are typically evaluated against labelled test sets with clear ground truth, generative models cannot be evaluated against the true data distribution directly. They are therefore assessed by comparing finite sets of real and generated samples drawn from complex high-dimensional distributions, which makes evaluation inherently difficult.

Diversity is a fundamental aspect of generative model evaluation, measuring how well generated samples cover the support and variation of the real data~\cite{sajjadi2018assessing,kynkaanniemiImprovedPrecisionRecall2019}. It affects both creative variation and the practical utility of generated samples in downstream applications, and has attracted increasing attention recently~\cite{astolfiConsistencydiversityrealismParetoFronts2024,sadatCADSUNLEASHINGDIVERSITY2024,meng2025training,trang2025discovering}. Diversity is often assessed indirectly through FID~\cite{heusel2017gans}, one of the most widely used evaluation metrics for image generation, as well as other set-based metrics such as IS~\cite{salimans2016improved}, KID~\cite{binkowski2018demystifying}, and the more recent CMMD~\cite{jayasumanaRethinkingFIDBetter2024}. However, these metrics collapse fidelity and diversity into a single number~\cite{naeemReliableFidelityDiversity,kynkaanniemiImprovedPrecisionRecall2019}. As a result, they do not indicate whether a score difference is caused by fidelity, diversity, or both. This limitation motivated precision--recall style evaluation~\cite{kynkaanniemiImprovedPrecisionRecall2019,naeemReliableFidelityDiversity}, where precision or density assesses sample quality, while recall or coverage assesses how well generated samples cover the real data distribution.

\begin{figure*}[t]
  \centering

  \begin{minipage}[t]{0.48\textwidth}
    \centering
    \includegraphics[width=\linewidth]{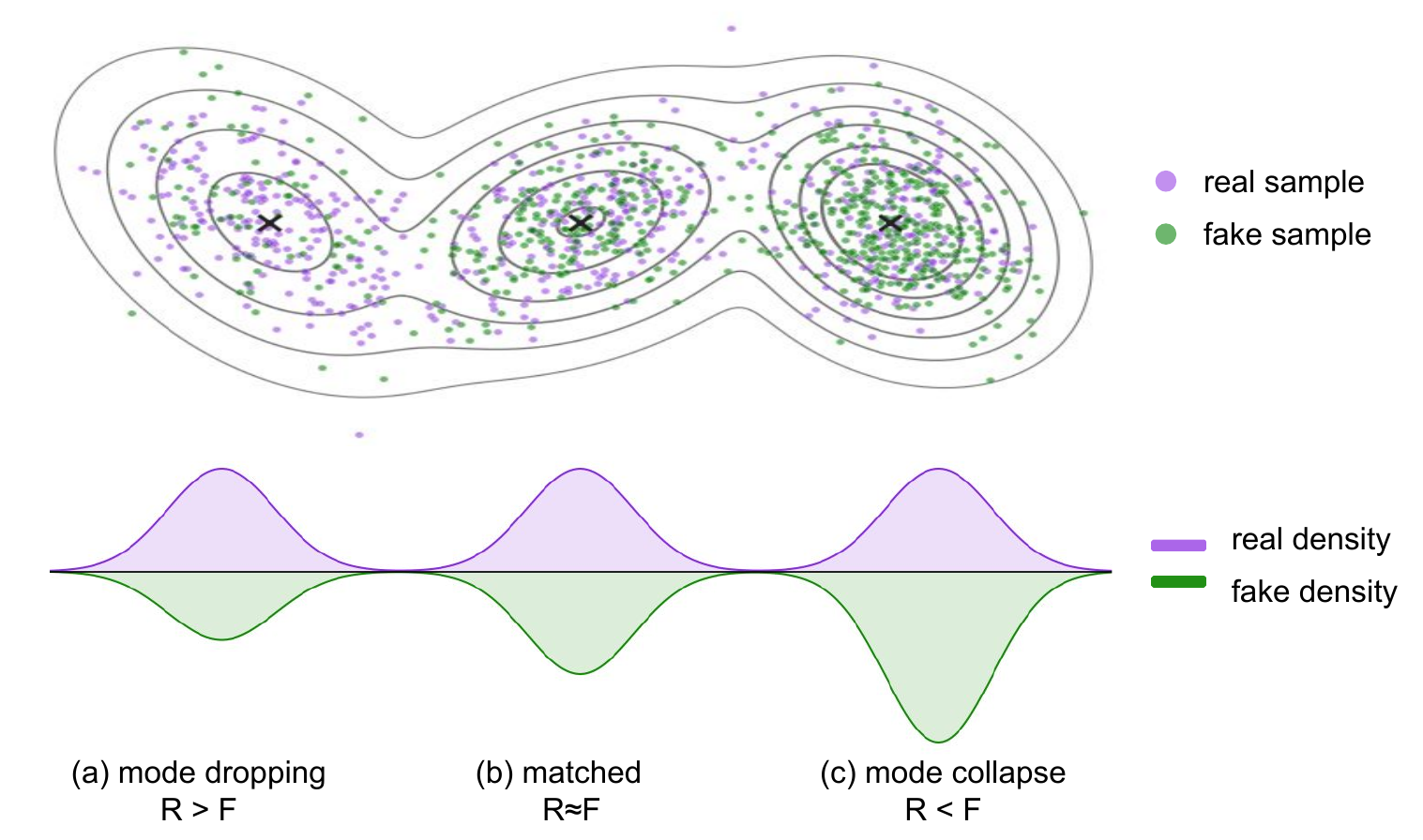}
\captionof{figure}{
Illustration of local density imbalance.
Rows show sample-level imbalance and the corresponding distribution-level view.
}

    \label{fig:cdensity}
  \end{minipage}
  \hfill
  \begin{minipage}[t]{0.48\textwidth}
    \centering
    \includegraphics[width=\linewidth]{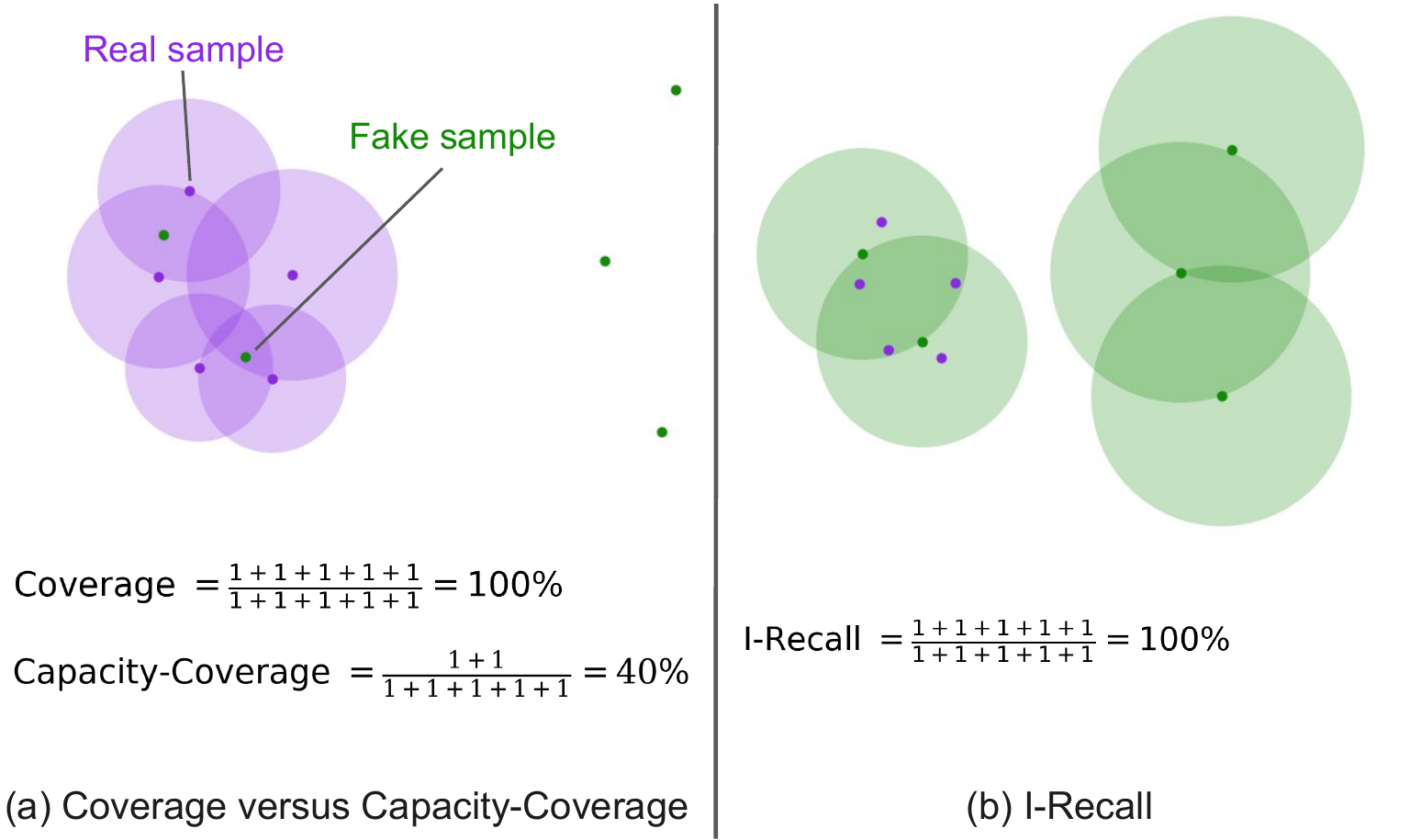}
    \captionof{figure}{Overview of metrics. An example illustrating the advantage of Capacity-Coverage over Coverage and I-Recall.}
    \label{fig:ccoverage}
  \end{minipage}

\end{figure*}

Recall is inherently a set-level property: its value depends not only on individual sample quality, but also on how generated samples are allocated across the real distribution. In our analysis, recall failures can arise from different allocation imbalances, especially mode dropping and mode collapse~\footnote{The terminology of mode dropping and mode collapse varies across prior work. Here, mode dropping denotes insufficient generated support in geometrically covered real-data regions, whereas mode collapse denotes local over-representation of generated samples. Some prior work also calls complete mode disappearance mode dropping. Our diagnostic excludes fully uncovered real samples, whose missing support instead lowers our capacity-aware coverage score.}, as illustrated in Fig.~\ref{fig:ccoverage}. For this reason, a useful recall metric should not only provide an overall score, but also report diagnostic scores that explain which coverage failure is present.

Although recall is intended to measure distributional coverage, most existing recall-style metrics simplify this problem into geometry-based coverage. In practice, they estimate local support through $k$NN-based neighbourhoods and then reduce recall to a single aggregate score~\cite{kynkaanniemiImprovedPrecisionRecall2019,naeemReliableFidelityDiversity,parkProbabilisticPrecisionRecall2023,kimTopPRRobustSupport}. This simplifying assumption weakens their diagnostic value and interpretability in two ways. First, a single recall score lacks information about which coverage failure types are present or how severe they are. For example, two generators may have similar recall scores while exhibiting different diversity failures, such as mode dropping and mode collapse. Second, geometry-based coverage can systematically overestimate recall. A small number of generated samples may cover many real samples in local neighbourhoods, even when the synthetic set is too sparse to provide balanced coverage~\cite{naeemReliableFidelityDiversity,parkProbabilisticPrecisionRecall2023,salvyEnhancedGenerativeModel2026,khayatkhoeiEmergentAsymmetryPrecision}. As a result, these metrics provide limited feedback for diagnosing coverage behaviour and improving generative models.

Based on this observation, we propose CADMA, a \textbf{C}apacity-\textbf{A}ware coverage \textbf{D}ecomposition framework for generative \textbf{M}odel \textbf{A}ssessment. CADMA evaluates distributional coverage through local capacity matching between real and generated samples, with matches restricted to neighbourhoods centred on real samples. This separates geometry-based coverage from capacity-supported coverage, allowing the evaluation to account for whether generated samples are allocated in accordance with the real-data distribution. CADMA then defines a capacity-aware coverage score together with diagnostic scores for mode dropping and mode collapse, providing both a more reliable coverage estimate and a more detailed diagnosis of diversity failure modes.

Our contributions are three-fold.
\begin{itemize}
    \item We identify a core limitation of existing recall-style metrics: they reduce distributional coverage to geometry-based coverage. This simplification weakens diagnostic interpretability by reporting only a single recall score and can assign high recall to poorly covered generative distributions.
    \item We introduce CADMA, a capacity-aware coverage evaluation framework based on local capacity matching between real and generated samples. This moves beyond geometry-based coverage by defining a capacity-aware coverage score and diagnostic scores for mode dropping and mode collapse.
    \item We validate CADMA on both toy and real-data benchmarks, showing that it provides more reliable distributional coverage estimates and finer-grained, interpretable diagnostics of diversity failures than standard recall-oriented metrics. Despite its additional diagnostic ability, CADMA adds only about 1\% computational overhead.

\end{itemize}

\section{Preliminaries}
\label{sec:preliminaries}
\subsection{Problem Setting}
\label{subsec:problem_setting}

Let $\mathcal{R}_0=\{x_i\}_{i=1}^N$ and $\mathcal{F}_0=\{y_j\}_{j=1}^M$ denote the real and generated (fake) sample sets. Direct statistical comparison in the raw input space is generally difficult for high-dimensional data such as images. Following standard practice in generative model evaluation~\cite{heusel2017gans,salimans2016improved,kynkaanniemiImprovedPrecisionRecall2019,naeemReliableFidelityDiversity,parkProbabilisticPrecisionRecall2023}, we map both sets into a feature space using a pretrained feature extractor $\phi:\mathcal{X}\to\mathbb{R}^d$:
\[
\mathcal{R} = \{r_i\}_{i=1}^N := \{\phi(x_i)\}_{i=1}^N,
\qquad
\mathcal{F} = \{f_j\}_{j=1}^M := \{\phi(y_j)\}_{j=1}^M.
\]
In the rest of the paper, all distances and neighbourhoods are computed in this feature space.

\subsection{Geometry-Based Coverage in \texorpdfstring{$k$}{k}-NN-Based Recall}
\label{subsec:knn_metrics}

To avoid ambiguity, we use \emph{I-Recall} and \emph{Coverage} to denote the specific metrics. Both are recall-style metrics that estimate how well generated samples cover the real distribution. We use lowercase recall and coverage for the general evaluation concepts, treating them as interchangeable in this context because both refer to the coverage of the real distribution by generated samples.

At the distribution level, recall measures the probability that a real sample lies in the support of the generated distribution~\cite{parkProbabilisticPrecisionRecall2023,salvyEnhancedGenerativeModel2026}:
\begin{equation}
\mathrm{Recall}(P,Q)
:=
\mathbb{E}_{X \sim P}
\left[
\mathbf{1}\{X \in \mathrm{supp}(Q)\}
\right],
\label{eq:distribution_recall}
\end{equation}
where $P$ and $Q$ denote the real and generated distributions, respectively, $\mathrm{supp}(\cdot)$ denotes distributional support, and $\mathbf{1}[\cdot]$ denotes the indicator function, which equals one when its argument is true and zero otherwise. Since the underlying distributions are unknown,
Eq.~\eqref{eq:distribution_recall} cannot be evaluated directly.
In practice, these distributions are represented empirically by the
finite real and generated feature sets $\mathcal{R}$ and
$\mathcal{F}$, respectively.

In practice, \emph{I-Recall}~\cite{kynkaanniemiImprovedPrecisionRecall2019} approximates the generated support by the union of hyperspheres $\bigcup_{j=1}^{M} B(f_j,\eta_j)$ in $\mathbb{R}^d$ centered at generated features $f_j$, where the radius $\eta_j := \mathrm{NND}_k(f_j)$ is the distance from $f_j$ to its $k$-th nearest neighbour in $\mathcal{F}\setminus\{f_j\}$:
\begin{equation}
\mathrm{I\text{-}Recall}
:=
\frac{1}{N}
\sum_{i=1}^{N}
\mathbf{1}\!\left[
r_i \in \bigcup_{j=1}^{M} B(f_j,\eta_j)
\right].
\label{eq:irecall_clean}
\end{equation}
This formulation measures whether each real feature $r_i$ lies inside at least one generated neighbourhood.

Symmetrically, the \emph{Coverage} metric~\cite{naeemReliableFidelityDiversity} uses hyperspheres centered at real features $r_i$ and measures the fraction of such hyperspheres that contain at least one generated feature:
\begin{equation}
\mathrm{Coverage}
:=
\frac{1}{N}
\sum_{i=1}^{N}
\mathbf{1}\!\biggl[
\exists j\ \mathrm{s.t.}\ f_j \in B(r_i,\rho_i)
\biggr],
\label{eq:coverage_clean}
\end{equation}
where $\rho_i := \mathrm{NND}_k(r_i)$ is the corresponding real-side radius.

\paragraph{Advantage of real-induced geometry.}
The advantage of the \emph{Coverage} metric over \emph{I-Recall} is that it defines support hyperspheres on real data rather than generated samples. This real-induced geometry is more robust, because the support construction is not distorted by abnormal or out-of-distribution generated samples, and more efficient, because the reference hyperspheres only need to be built once and can be reused across generative models~\cite{naeemReliableFidelityDiversity}. In contrast, \emph{I-Recall} constructs hyperspheres from generated samples, so its estimated support must be recomputed for each model and can be sensitive to generated outliers~\cite{naeemReliableFidelityDiversity,parkProbabilisticPrecisionRecall2023,kimTopPRRobustSupport,alaaHowFaithfulYour,cheemaPrecisionRecallCover}.

\paragraph{Limitation of geometry-based coverage.}
Nevertheless, this real-induced construction still inherits a broader limitation of recall-style metrics: both \emph{I-Recall} and the \emph{Coverage} metric reduce distributional coverage to geometry-based coverage. As a result, they report a single summary score with limited diagnostic interpretability and cannot reveal the degree of different coverage failures, such as mode collapse and mode dropping. In addition, geometry-based coverage only checks whether each hypersphere contains at least one sample. Therefore, a small number of generated samples may make many real regions appear covered, leading to overly optimistic coverage estimates~\cite{parkProbabilisticPrecisionRecall2023}.

Subsequent works have proposed many alternatives and refinements to recall-style evaluation, with most relying on $k$NN-based neighbourhood constructions~\cite{parkProbabilisticPrecisionRecall2023,kimTopPRRobustSupport,alaaHowFaithfulYour,cheemaPrecisionRecallCover,khayatkhoeiEmergentAsymmetryPrecision,salvyEnhancedGenerativeModel2026}. However, these methods still largely inherit a common geometry-based view of coverage. Recent sanity-check benchmarks further show that existing recall-style metrics exhibit important failure cases and unreliable estimates~\cite{raisaPositionAllCurrent2025}.

CADMA follows the real-induced construction of the \emph{Coverage} metric and inherits its robustness and efficiency, while addressing the limitations of geometry-based coverage through capacity-aware real--fake matching. This matching provides a more fine-grained and interpretable diagnosis of coverage failures, including mode collapse and mode dropping, while reducing overly optimistic coverage estimates.

\section{Method}
\label{sec:method}

Motivated by the limitations above, we propose CADMA, a recall-style metric that reports the degrees of mode dropping and mode collapse while reducing overly optimistic coverage estimates. Overly optimistic estimates can arise when a small number of generated samples make a large portion of the real distribution appear covered. CADMA addresses this issue through \emph{local capacity matching}, which prevents the same generated sample from being repeatedly used to support many real samples. We first define this capacity-aware coverage score, and then show how the same matching structure induces diagnostic scores for mode dropping and mode collapse.

\subsection{Local Capacity Matching}
\label{subsec:local_capacity_matching}

We first define the local coverage relation using the same real-induced neighbourhoods as the~\emph{Coverage} metric~\cite{naeemReliableFidelityDiversity}. Let $\rho_i=\mathrm{NND}_k(r_i)$ (Eq.~\ref{eq:coverage_clean}) denote the distance from real feature $r_i$ to its $k$-th nearest neighbour in $\mathcal{R}\setminus\{r_i\}$. We define a binary matrix $A$ by
\begin{equation}
A_{ij}
=
\mathbf{1}\!\left[
\|r_i-f_j\|_2 \le \rho_i
\right],
\qquad
A \in \{0,1\}^{N\times M}.
\label{eq:adjacency_matrix}
\end{equation}
Here, $A_{ij}=1$ means that generated feature $f_j$ falls inside the real-induced hypersphere of real feature $r_i$. The matrix $A$ records all local real--generated relations: one real hypersphere may contain multiple generated features, and one generated feature may fall inside multiple real hyperspheres. Equivalently, some rows may satisfy $\sum_{j=1}^{M}A_{ij}>1$, and some columns may satisfy $\sum_{i=1}^{N}A_{ij}>1$.

CADMA therefore introduces \emph{local capacity matching}, which extracts a matching matrix $Z$ from the admissible relations in $A$. It is local because matching can only use real-induced neighbourhood relations,
and capacity-constrained because each real or generated feature can be matched at most once:
\begin{equation}
\begin{aligned}
Z_{ij}&\le A_{ij}, &&\forall i,j, \qquad
\sum_{j=1}^{M} Z_{ij}\le 1, &&\forall i, \qquad
\sum_{i=1}^{N} Z_{ij}\le 1, &&\forall j.
\end{aligned}
\label{eq:matching_constraints}
\end{equation}

The matrix $A$ defines an unweighted bipartite graph, with real and generated features forming its two vertex sets and $A_{ij}=1$ defining an edge between $r_i$ and $f_j$. Multiple matching matrices $Z$ may satisfy Eq.~\eqref{eq:matching_constraints}. Among these feasible matchings, CADMA selects one with maximum cardinality:
\begin{equation}
Z^{\star}
\in
\arg\max_{Z\in\{0,1\}^{N\times M}}
\sum_{i=1}^{N}\sum_{j=1}^{M} Z_{ij}
\quad
\text{s.t. Eq.~\eqref{eq:matching_constraints}}.
\label{eq:max_matching_matrix}
\end{equation}

We use the resulting matching size $N^{\mathrm{mat}}=\sum_{i=1}^{N}\sum_{j=1}^{M}Z^\star_{ij}$. We compute a maximum matching using the Hopcroft--Karp algorithm~\cite{hopcroft1973n}, a standard exact solver for this problem. Any exact maximum-matching solver yields the same value of $N^{\mathrm{mat}}$.

\subsection{Capacity-Aware Coverage and Diagnostic Scores}
\label{subsec:coverage_diagnostics}

Local capacity matching yields a capacity-aware coverage score and two diagnostic scores for mode dropping and mode collapse. Following common practice in set-based generative evaluation, we use equally sized real and generated sample sets ($N=M$), so real-side and generated-side excesses are comparable.

The \emph{Coverage} metric in Eq.~\eqref{eq:coverage_clean} counts real samples whose neighbourhood contains at least one generated sample. In our notation, let
\[
u_i=\mathbf{1}\!\left[\sum_{j=1}^{M}A_{ij}\ge 1\right],
\qquad
N^{\mathrm{cov}}=\sum_{i=1}^{N}u_i.
\]
Since $A$ is defined using the same real-side radius $\rho_i$ as in Eq.~\eqref{eq:coverage_clean}, the \emph{Coverage} metric can be written as
\[
\mathrm{Coverage}
=
\frac{N^{\mathrm{cov}}}{N}.
\]

CADMA instead defines capacity-aware coverage from the maximum matching size under local capacity matching:
\begin{equation}
\mathcal{C}_{\mathrm{cap}}
:=
\frac{N^{\mathrm{mat}}}{N}.
\label{eq:capacity_coverage}
\end{equation}

Mode dropping is reflected in local regions where real samples outnumber the available generated samples. In geometry-based coverage, such regions may still be counted as covered as long as each real neighbourhood contains at least one generated sample. After local capacity matching, however, only $N^{\mathrm{mat}}$ of the $N^{\mathrm{cov}}$ covered real samples can be paired with generated samples. Thus, $N^{\mathrm{cov}}-N^{\mathrm{mat}}$ counts geometrically covered real samples that remain unmatched due to a lack of distinct generated counterparts. Normalizing this quantity by the total number of real samples gives the mode dropping score:
\begin{equation}
\mathcal{D}_{\mathrm{drop}}
:=
\mathrm{Coverage}-\mathcal{C}_{\mathrm{cap}}
=
\frac{N^{\mathrm{cov}}-N^{\mathrm{mat}}}{N}.
\label{eq:mode_dropping_score}
\end{equation}

Equivalently, the \emph{Coverage} metric decomposes as $\mathrm{Coverage}=\mathcal{C}_{\mathrm{cap}}+\mathcal{D}_{\mathrm{drop}}$. This decomposition explains why geometry-based coverage can be overly optimistic: it combines effective coverage $\mathcal{C}_{\mathrm{cap}}$ with a mode-dropping component $\mathcal{D}_{\mathrm{drop}}$. By separating this mode-dropping component, CADMA obtains a more reliable capacity-aware coverage estimate.

Symmetrically, mode collapse appears as local regions where generated samples outnumber the available real samples. We define the generated-side covered count as
\[
v_j=\mathbf{1}\!\left[\sum_{i=1}^{N}A_{ij}\ge 1\right],
\qquad
M^{\mathrm{cov}}=\sum_{j=1}^{M}v_j.
\]
Here, $M^{\mathrm{cov}}$ counts generated samples that are locally covered by the real-induced neighbourhoods. Under local capacity matching, only $N^{\mathrm{mat}}$ of these $M^{\mathrm{cov}}$ generated samples can be paired with distinct real samples. The remaining $M^{\mathrm{cov}}-N^{\mathrm{mat}}$ therefore measures the excess covered generated samples that lack real counterparts. Normalizing this quantity by the total number of real samples gives the mode-collapse score:

\begin{equation}
\mathcal{D}_{\mathrm{collapse}}
:=
\frac{M^{\mathrm{cov}}-N^{\mathrm{mat}}}{N}.
\label{eq:mode_collapse_score}
\end{equation}

\subsection{Radius Regularization}
\label{subsec:radius_clipping}

Under the raw radius choice $\varepsilon_i=\rho_i$, the geometry-based score reduces to the \emph{Coverage} metric. However, raw $k$NN radii can become excessively large in sparse regions or around outliers, making the induced support overly permissive~\cite{naeemReliableFidelityDiversity,parkProbabilisticPrecisionRecall2023,khayatkhoeiEmergentAsymmetryPrecision}. To stabilize the neighbourhood geometry, we cap each local radius by the global mean real-side radius, $\bar{\rho}
=
\frac{1}{N}\sum_{i=1}^{N}\rho_i,
$
and define
\begin{equation}
\varepsilon_i
=
\min(\rho_i,\bar{\rho}).
\label{eq:min_mean_radius}
\end{equation}

This regularization is designed to reduce the influence of outlier-induced large neighbourhoods while preserving local radii in dense regions. We empirically verify its effect in the outlier ablation in Fig.~\ref{fig:ablation_radius}.

\begin{figure*}[tbp]
    \centering
    \includegraphics[width=1\linewidth]{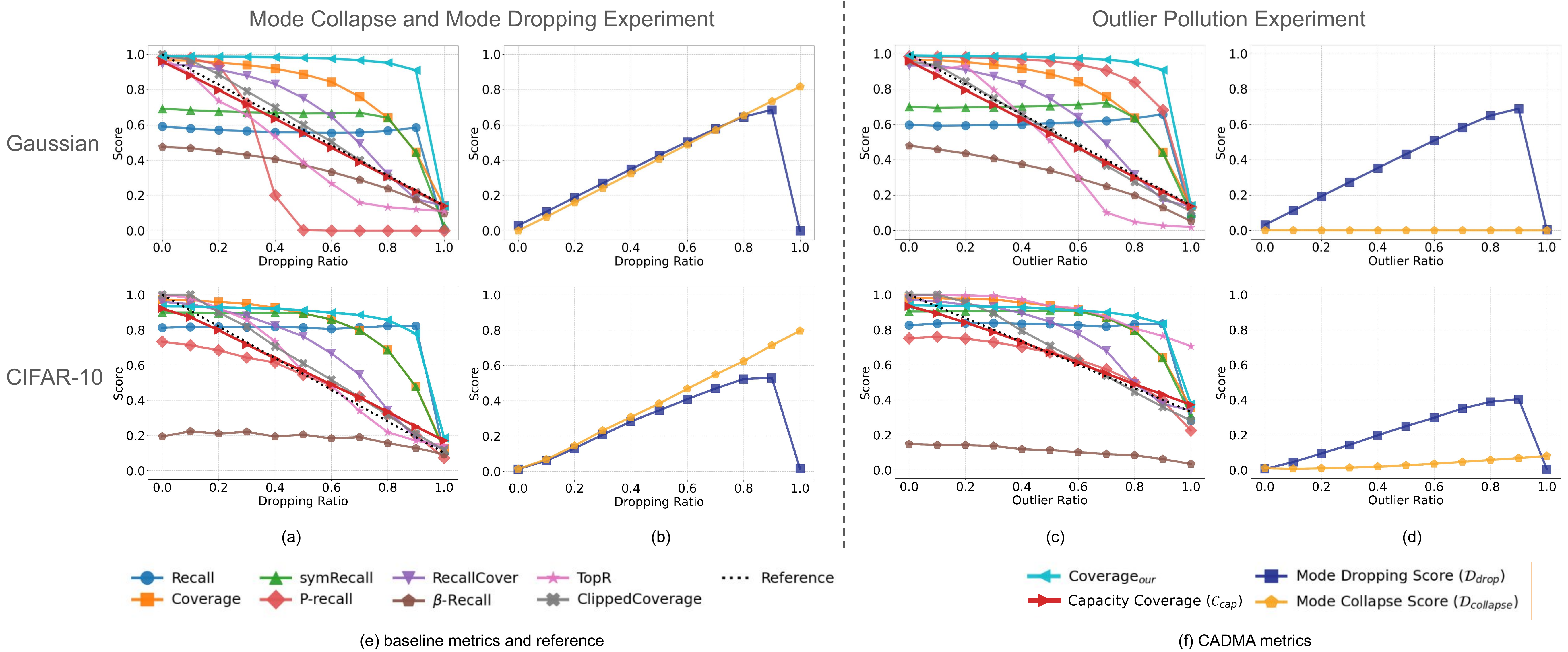}
\caption{
Main experiments on Gaussian and CIFAR-10 mode dropping, mode collapse, and outlier pollution.
We compare the proposed capacity-aware coverage with existing baselines and report CADMA diagnostic scores for mode dropping and mode collapse.
}

    \label{fig:mode_dropping_gaussian_cifar}
\end{figure*}

\section{Experiments}
\label{sec:experiments}

\paragraph{Experimental Setup}
We evaluate CADMA against a broad collection of existing recall-style metrics, including $\beta$-Recall, TopR~\cite{kimTopPRRobustSupport}, I-Recall~\cite{kynkaanniemiImprovedPrecisionRecall2019}, Coverage~\cite{naeemReliableFidelityDiversity}, symRecall~\cite{khayatkhoeiEmergentAsymmetryPrecision}, P-recall~\cite{parkProbabilisticPrecisionRecall2023}, Recall Cover~\cite{cheemaPrecisionRecallCover}, and ClippedCoverage~\cite{salvyEnhancedGenerativeModel2026}. For method-specific hyperparameters, we follow the original recommendations. In particular, we set $k'=3$ and $C=3$ for Precision Recall Cover, $k=3$ for I-Recall, and $k=4$ for P-recall. For CADMA, we use $k=70$ unless otherwise stated.
An ablation of this choice is given in Sec.~\ref{subsec:exp_ablation}. All experiments are run on a single NVIDIA H100 GPU with 80GB memory unless otherwise stated. For image-based evaluations, we extract features using DINOv2-Large (ViT-L/14)~\cite{oquab2023dinov2}, following recent practice~\cite{salvyEnhancedGenerativeModel2026,foxUnifyingInformationtheoreticPerspective} and prior evidence that this representation space is more semantically meaningful and better aligned with human perception for generative image evaluation~\cite{steinExposingFlawsGenerative}.

Evaluating generative model metrics is inherently difficult because ground-truth metric values are typically unavailable. Following prior work, we assess CADMA through controlled sanity checks on toy and real-world data where the expected metric behaviour can be clearly specified. Additional evaluations of existing generators on ImageNet-256,
FFHQ, and LSUN Bedroom are provided in
Appendix~\ref{app:models}.
Theoretical complexity analysis and runtime and memory measurements
are presented in Appendix~\ref{app:scalability}.

\begin{table*}[t]
\centering
\caption{
Mean absolute error. Best and second-best results are bolded and underlined.
}

\label{tab:mae_error_refer}
\resizebox{\textwidth}{!}{
\begin{tabular}{lccccc}
\toprule
\multirow{2}{*}{Method}
& \multicolumn{2}{c}{Gaussian}
& \multicolumn{2}{c}{CIFAR-10}
& \multirow{2}{*}{Avg.} \\
\cmidrule(lr){2-3} \cmidrule(lr){4-5}
& Mode Collapse ($\downarrow$) & Outlier Pollution ($\downarrow$)
& Mode Collapse ($\downarrow$) & Outlier Pollution ($\downarrow$)
& \\
\midrule
Recall~\cite{kynkaanniemiImprovedPrecisionRecall2019}
                & 0.206 & 0.214 & 0.252 & 0.179 & 0.213 \\
Coverage~\cite{naeemReliableFidelityDiversity}
                & 0.204 & 0.203 & 0.239 & 0.189 & 0.209 \\
symRecall~\cite{khayatkhoeiEmergentAsymmetryPrecision}
                & 0.180 & 0.183 & 0.227 & 0.175 & 0.191 \\
P-recall~\cite{parkProbabilisticPrecisionRecall2023}
                & 0.255 & 0.283 & 0.080 & 0.079 & 0.174 \\
RecallCover~\cite{cheemaPrecisionRecallCover}
                & 0.088 & 0.085 & 0.108 & 0.090 & 0.093 \\
$\beta$-Recall~\cite{alaaHowFaithfulYour}
                & 0.230 & 0.260 & 0.368 & 0.560 & 0.355 \\
TopR~\cite{kimTopPRRobustSupport}
                & 0.119 & 0.127 & 0.054 & 0.240 & 0.135 \\
ClippedCoverage~\cite{salvyEnhancedGenerativeModel2026}
                & \underline{0.025} & \textbf{0.020} & \underline{0.051} & \underline{0.045} & \underline{0.035} \\
\midrule
CADMA ($\mathcal{C}_{\mathrm{cap}}$)
                & \textbf{0.019} & \underline{0.023} & \textbf{0.039} & \textbf{0.025} & \textbf{0.027} \\
\bottomrule
\end{tabular}
}
\end{table*}

\subsection{Mode Dropping and Mode Collapse}
\label{subsec:exp_mode_dropping}

We evaluate whether CADMA can diagnose mode dropping and mode collapse. Following prior mode-dropping settings~\cite{salvyEnhancedGenerativeModel2026,naeemReliableFidelityDiversity,kimTopPRRobustSupport}, we manually remove samples from selected modes. In the density-shift setting, removed samples are reassigned to remaining modes, producing both mode dropping and mode collapse. In the outlier-pollution setting, removed samples are replaced by outliers, testing whether CADMA separates missing coverage from generated samples outside the real support.

\subsubsection{Density Shift in Gaussian Mixtures}

We first evaluate controlled density shift on a seven-mode Gaussian mixture. Generated samples are progressively moved from six modes into the remaining mode. Consequently, the reference coverage decreases linearly from $1$ to $1/7$. Samples are removed from all six modes simultaneously, rather than eliminating entire modes sequentially (see Fig.~\ref{fig:supp_distribution_settings}). Thus, the affected modes remain present at intermediate stages but become progressively under-represented, creating local imbalance between the real and generated distributions.

As shown in Fig.~\ref{fig:mode_dropping_gaussian_cifar}(a), $\mathcal{C}_{\mathrm{cap}}$ closely follows this reference trend, while most baselines deviate substantially. Several baselines still assign high recall scores at large dropping ratios, indicating optimistic estimates under severe density shift. We also report $\mathrm{Coverage}_{\mathrm{our}}$ as a geometry-based reference, computed using the same neighbourhoods as $\mathcal{C}_{\mathrm{cap}}$ but without capacity matching.

ClippedCoverage is the closest baseline to the reference trend, but it remains a single-score metric. CADMA additionally reports mode-dropping and mode-collapse diagnostics in Fig.~\ref{fig:mode_dropping_gaussian_cifar}(b). As the dropping ratio increases, $\mathcal{D}_{\mathrm{collapse}}$ rises because generated samples concentrate in the remaining mode, while $\mathcal{D}_{\mathrm{drop}}$ initially rises because affected modes receive insufficient generated support. At the final stage, $\mathcal{D}_{\mathrm{drop}}$ drops because these regions become fully uncovered rather than locally under-supported, which is consistent with its definition.

\subsubsection{Density Shift in CIFAR-10}

We further evaluate the same density-shift protocol in the CIFAR-10 feature space, following the real-data setting in ClippedCoverage~\cite{salvyEnhancedGenerativeModel2026}. We form two equal-sized sets from CIFAR-10 real data, treating one as the real set and the other as the generated set, using all 10 classes with 2500 samples per class in each set. We then simultaneously drop samples from nine classes and reassign them to the remaining class, creating mode dropping together with mode collapse. The reference recall trend decreases linearly from $1$ to $1/10$ as the dropping ratio increases from $0\%$ to $90\%$.

As shown in Fig.~\ref{fig:mode_dropping_gaussian_cifar}(a), the Gaussian conclusions carry over to real image features: $\mathcal{C}_{\mathrm{cap}}$ tracks the reference trend more faithfully than most baselines, while several existing metrics remain overly high under severe dropping. Fig.~\ref{fig:mode_dropping_gaussian_cifar}(b) further shows that CADMA provides diagnostic information beyond the recall score. As the dropping ratio increases, $\mathcal{D}_{\mathrm{collapse}}$ reflects the increasing concentration of generated samples in the remaining class, while $\mathcal{D}_{\mathrm{drop}}$ captures the loss of effective generated support in affected classes.

\begin{figure*}[t]
  \centering

  \begin{minipage}[t]{0.64\textwidth}
    \centering
    \includegraphics[width=\linewidth]{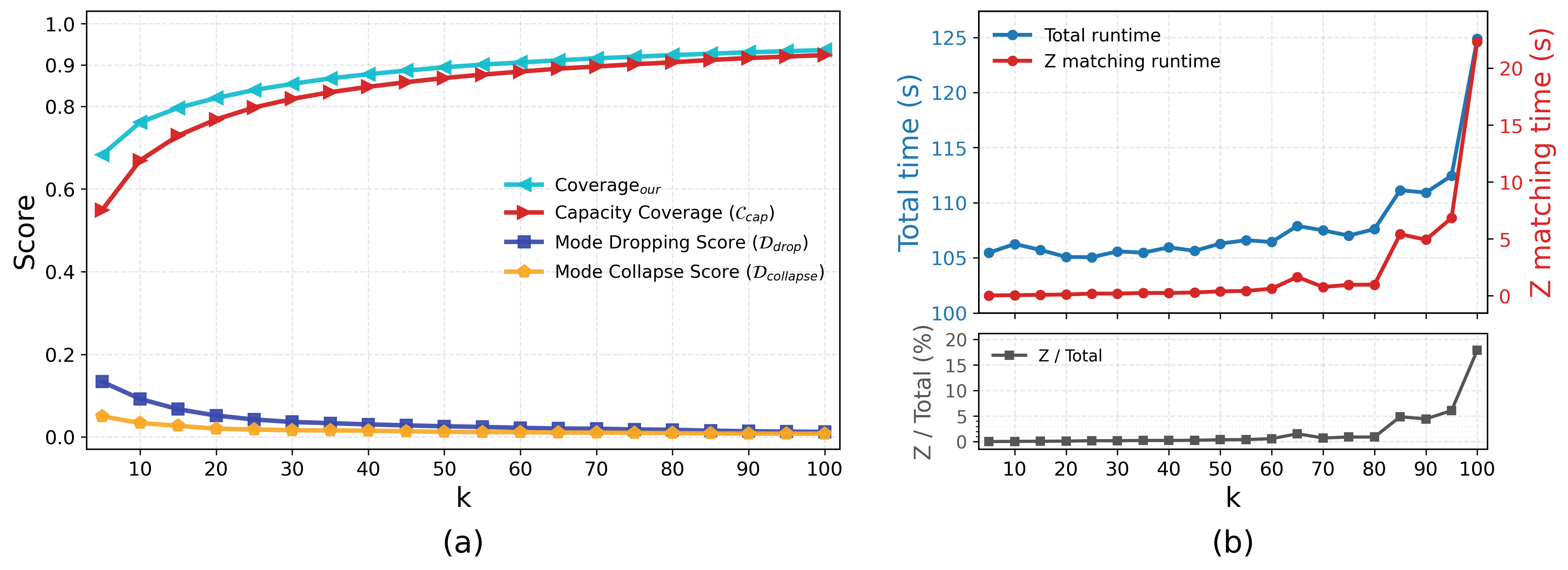}
    \captionof{figure}{
    Ablation on the hyperparameter $k$, showing the trade-off between CADMA scores in (a) and runtime in (b).
    }
    \label{fig:ablation_k}
  \end{minipage}
  \hspace{0.015\textwidth}
  \begin{minipage}[t]{0.32\textwidth}
    \centering
    \includegraphics[width=\linewidth]{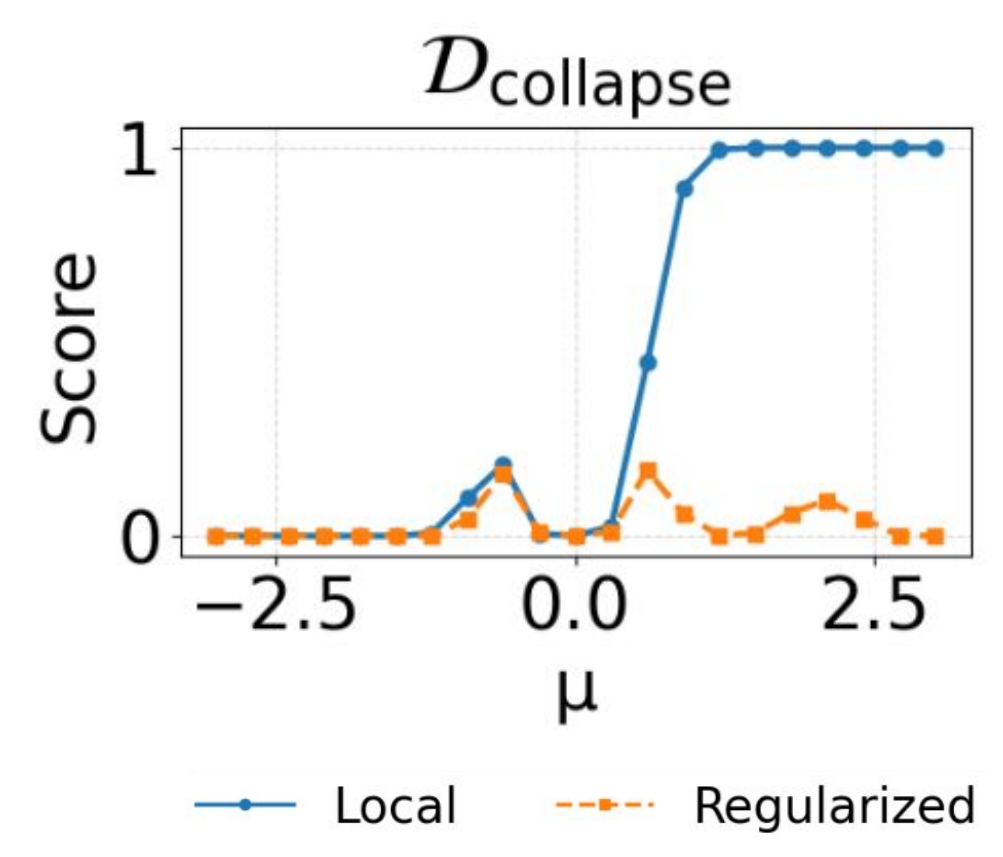}
    \captionof{figure}{
    Ablation on the radius regularization strategy.
    }
    \label{fig:ablation_radius}
  \end{minipage}
  \hspace{0.005\textwidth}

\end{figure*}

\subsection{Robustness under Outlier Pollution}
\label{subsec:exp_outlier}

\paragraph{Asymmetric Gaussian translation with outliers.}Following prior work on outlier robustness and asymmetric evaluation~\cite{naeemReliableFidelityDiversity,parkProbabilisticPrecisionRecall2023,khayatkhoeiEmergentAsymmetryPrecision,kynkaanniemiImprovedPrecisionRecall2019}, we compare $X \sim \mathcal{N}(0,I)$ and $Y \sim \mathcal{N}(\mu,I)$ for $\mu \in [-3,3]$, with one real outlier from $\mathcal{N}(2,I)$ and one generated outlier from $\mathcal{N}(-2,I)$. Since the translation is symmetric around $\mu=0$, a metric robust to outlier pollution should retain
approximate symmetry.

As shown in Fig.~\ref{fig:gaussian_translation_outlier}, \emph{I-Recall} becomes strongly asymmetric under outlier pollution, reflecting sensitivity to generated-sample-induced neighbourhood inflation. CADMA remains more symmetric in both the capacity-aware coverage comparison in Fig.~\ref{fig:gaussian_translation_outlier}(a) and the diagnostic scores in Fig.~\ref{fig:gaussian_translation_outlier}(b). Although the mode-collapse score is not perfectly symmetric, radius regularization makes it less sensitive to outlier-induced fluctuations, as shown in Fig.~\ref{fig:ablation_radius}.

\paragraph{CIFAR-10 outlier pollution.}
To test robustness, sensitivity, and interpretability under outlier pollution, we use six CIFAR-10 categories as in-distribution support and treat the remaining four as outliers. We progressively replace generated samples from the in-distribution categories with outlier-category samples, as shown in Fig.~\ref{fig:mode_dropping_gaussian_cifar}(c,d). Since the proportion of valid generated samples decreases, an ideal recall-oriented metric should decrease linearly. At the final stage, only two in-distribution categories remain covered, so the ideal recall decreases from $1$ to $2/6 \approx 0.33$.

Among the coverage metrics, only $\mathcal{C}_{\mathrm{cap}}$ and ClippedCoverage exhibit this desired linear degradation. CADMA further provides diagnostic information beyond this recall trend. As valid generated samples are replaced, $\mathcal{D}_{\mathrm{drop}}$ initially increases because real samples lose effective generated support. In contrast, $\mathcal{D}_{\mathrm{collapse}}$ remains low because the injected outliers do not belong to the real distribution and therefore should not be interpreted as mode collapse. This differs from the density-shift setting in Fig.~\ref{fig:mode_dropping_gaussian_cifar}(a,b), where generated samples concentrate on the remaining modes. The observed pattern in Fig.~\ref{fig:mode_dropping_gaussian_cifar}(d), with initially increasing $\mathcal{D}_{\mathrm{drop}}$ and nearly unchanged $\mathcal{D}_{\mathrm{collapse}}$, supports the interpretability of CADMA's diagnostic scores.

\subsection{Quantitative results}
\label{subsec:reference_error}

\paragraph{MAE Error to the Reference Trend}
We also report the mean absolute error to the reference trend on Gaussian and CIFAR-10 experiments in Table~\ref{tab:mae_error_refer}. The reference trends are labelled ``Reference'' in Fig.~\ref{fig:mode_dropping_gaussian_cifar}. CADMA's capacity-aware coverage $\mathcal{C}_{\mathrm{cap}}$ achieves the lowest average error across all settings, with ClippedCoverage as the closest baseline.

This comparison is particularly strong because linear score degradation is a central design goal of ClippedCoverage: its main contribution is to calibrate coverage scores so that they follow an interpretable linear trend under controlled degradation. Nevertheless, CADMA achieves a lower average error than ClippedCoverage, showing that capacity-constrained matching provides a more faithful estimate of the effective recall trend even on the criterion where ClippedCoverage is explicitly designed to perform well.

\paragraph{Runtime of Matching}
Fig.~\ref{fig:ablation_k}(b) shows that the matching step is lightweight. Both the total runtime and the runtime for solving $Z^\star$ increase only mildly as $k$ grows. For $k \le 80$, the matching step accounts for about 1\% of the total runtime. Further theoretical complexity analysis and runtime and memory measurements across sample sizes are provided in Appendix~\ref{app:scalability}.

\subsection{Ablation Studies}
\label{subsec:exp_ablation}

\paragraph{Effect of radius regularization.}
Fig.~\ref{fig:ablation_radius} compares raw and regularized radii in the Gaussian translation setting with outliers. Raw radii lead to a noticeably asymmetric curve, indicating sensitivity to outlier-induced neighbourhood inflation. Radius regularization produces a more symmetric trend, consistent with the underlying translation, so we use it throughout CADMA.

\paragraph{Effect of neighbourhood size $k$.}
Fig.~\ref{fig:ablation_k} evaluates the effect of $k$ on CIFAR-10 feature sets with matched class proportions. Since the real and generated sets are distributionally aligned, the ideal recall should be close to 1. Small values of $k$ define overly restrictive neighbourhoods, making local balance difficult to estimate and lowering $\mathcal{C}_{\mathrm{cap}}$. As $k$ increases, more admissible matches become available, but the matching graph also becomes denser and runtime increases. Importantly, both performance and runtime remain stable over a broad range, especially for $40 \le k \le 80$. This suggests that CADMA is not sensitive to precise tuning of $k$, reducing the burden of hyperparameter selection. We use $k=70$ unless otherwise stated.

\begin{figure*}[t]
  \centering

  \begin{minipage}[t]{0.59\textwidth}
    \centering
    \includegraphics[width=\linewidth]{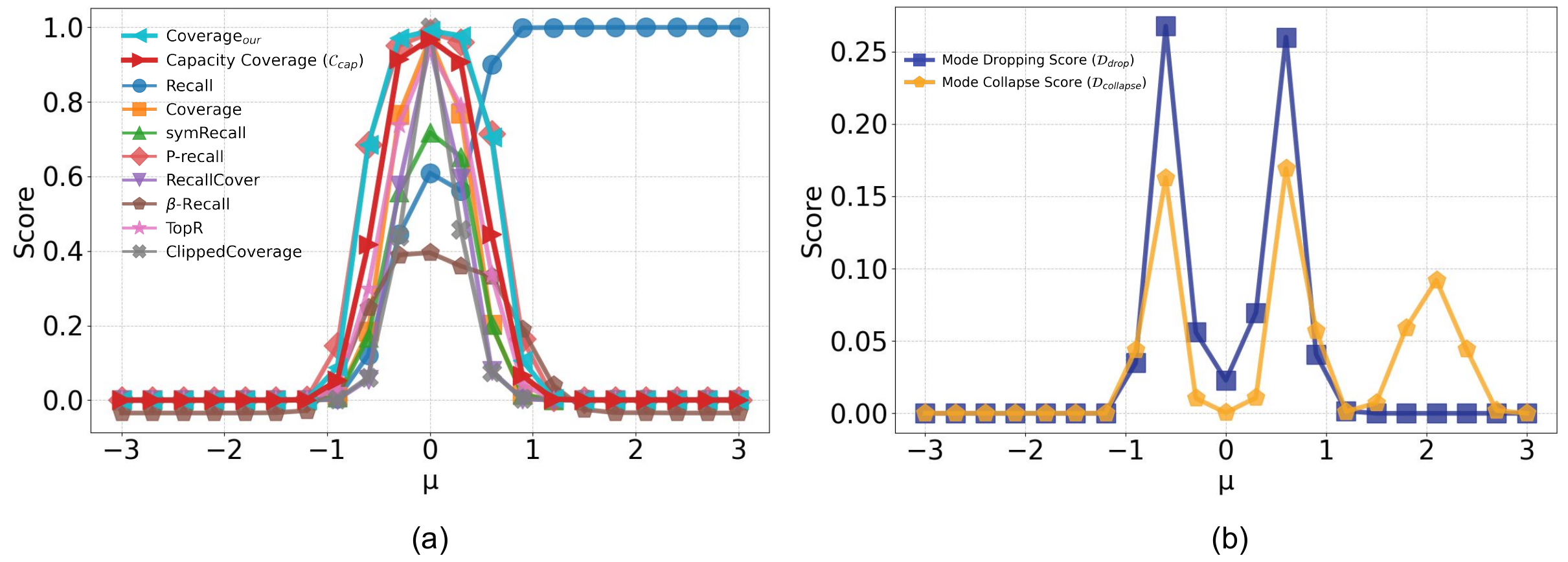}
    \captionof{figure}{
Translation and outlier experiments on Gaussian data.
CADMA remains stable and exhibits symmetric behaviour under distribution shifts.
}

    \label{fig:gaussian_translation_outlier}
  \end{minipage}
  \hspace{0.02\textwidth}
  \begin{minipage}[t]{0.36\textwidth}
    \centering
    \includegraphics[width=\linewidth]{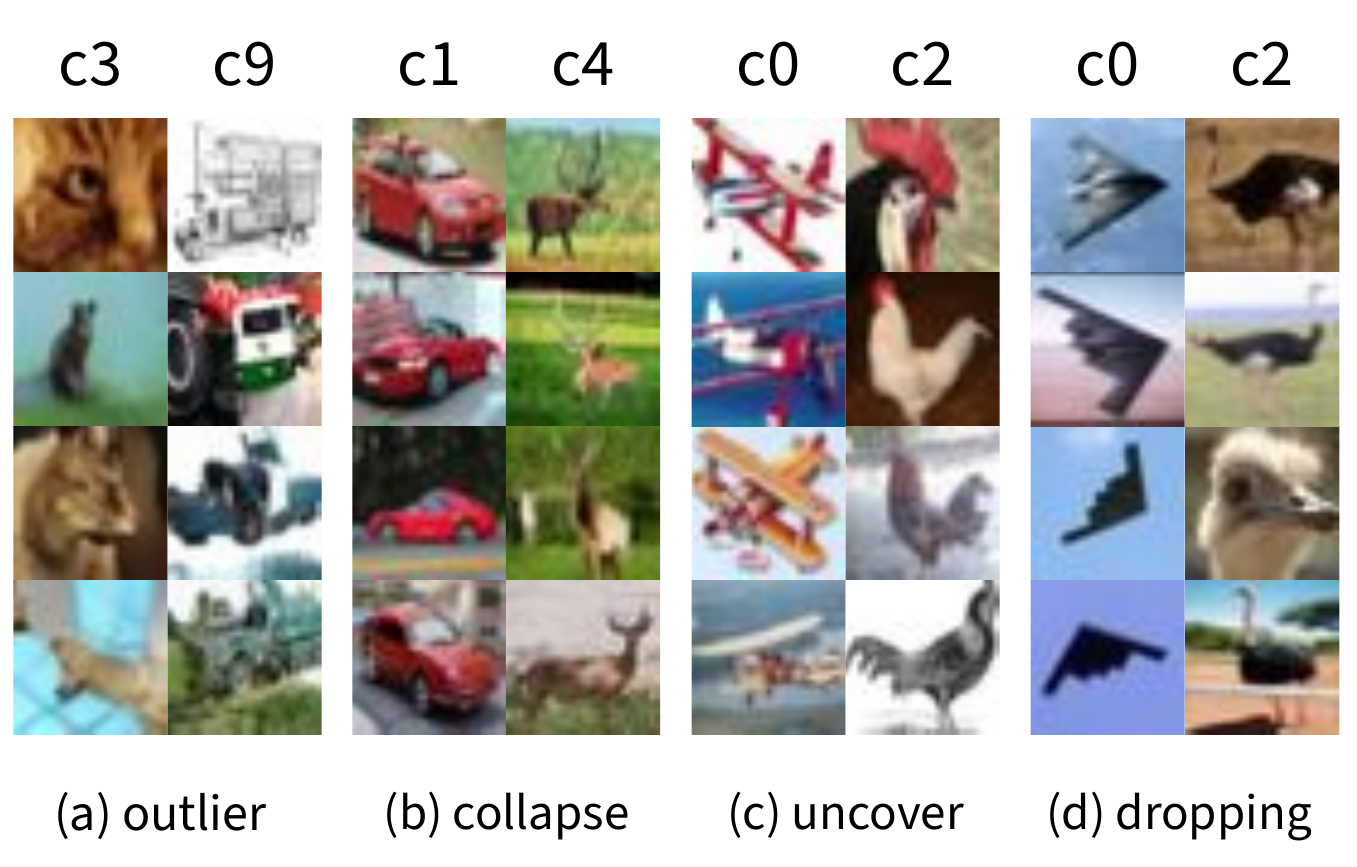}
    \captionof{figure}{
CIFAR-10 diagnostic visualization by CADMA.
The labels c0--c9 indicate CIFAR-10 categories.
}

    \label{fig:cifar_diagnosis}
  \end{minipage}

\end{figure*}

\paragraph{Visualization diagnosis and practical value.}
For this visualization, we evaluate CADMA using 25,000 images generated by the publicly released StyleGAN-XL model~\cite{sauer2022stylegan} across all 10 CIFAR-10 classes. Fig.~\ref{fig:cifar_diagnosis} visualizes representative CIFAR-10 samples from four CADMA regions: mode dropping, mode collapse, uncovered real samples, and generated outliers. These regions provide sample-level diagnostic information about model behaviour. For example, the mode-collapse region shows that the generator tends to over-produce certain patterns, such as red cars. The uncovered region reveals visual concepts that the model fails to generate, such as old-style aircraft and roosters. In contrast, the mode-dropping region contains concepts that the model can generate but only weakly represents, such as bat-like aircraft and ostrich-like birds.

\section{Related Work}

\paragraph{Single-score generative evaluation.} Classical distribution-level metrics include Inception Score (IS)~\citet{barratt2018note,salimans2016improved}, Kernel Inception Distance (KID)~\cite{binkowski2018demystifying,xu2018empirical}, Fr\'echet Inception Distance (FID)~\cite{heusel2017gans}, and more recently CMMD~\cite{jayasumanaRethinkingFIDBetter2024}. Recent text-to-image evaluation has also introduced sample-level alignment and preference metrics such as CLIPScore~\cite{hessel2021clipscore}, PickScore~\cite{kirstainPickaPicOpenDataset2023}, ImageReward~\cite{xuImageRewardLearningEvaluating2023}, TIFA~\cite{hu2023tifa}, GenEval~\cite{ghosh2023geneval}, and VQAScore~\cite{lin2024evaluating}. In this work, we focus on distribution-level evaluation. Sample-level metrics mainly assess the quality or alignment of individual images, but they do not directly evaluate how well a generated set matches the real data distribution as a whole. They are therefore complementary to the distribution-level recall evaluation considered here.

Among distribution-level metrics, FID has become the dominant benchmark in image generation~\cite{salimans2016improved,heusel2017gans,binkowski2018demystifying,rombachHighResolutionImageSynthesis2022}. More recent distances such as CMMD further improve this line by relaxing the Gaussian feature assumption through an MMD-based formulation. However, both FID and CMMD summarise the comparison between two distributions into a single number, failing to separate two important aspects of generative model quality: fidelity and diversity~\cite{sajjadi2018assessing}. As a result, when one model achieves a better FID- or CMMD-style score than another, it is unclear whether the difference is driven by fidelity, diversity, or both.

\paragraph{Fidelity--diversity decomposition.}
Motivated by the observation that scalar metrics such as FID cannot distinguish fidelity from diversity, \citet{sajjadi2018assessing} introduced the original precision--recall framework for generative models. In this line of work, precision is associated with fidelity and measures the fraction of generated samples that are realistic, whereas recall is associated with diversity and measures how well generated samples cover the variation of the real data~\cite{salvyEnhancedGenerativeModel2026,naeemReliableFidelityDiversity,kynkaanniemiImprovedPrecisionRecall2019}. \citet{kynkaanniemiImprovedPrecisionRecall2019} later proposed Improved Precision and Recall, replacing clustering-based support estimation with a $k$NN-based construction. This made the framework easier to compute, but also introduced sensitivity to outliers in the construction of local neighbourhoods. \citet{naeemReliableFidelityDiversity} proposed Density and Coverage, and \citet{parkProbabilisticPrecisionRecall2023} introduced a probabilistic formulation of precision and recall to improve the robustness of the $k$NN-based construction to outliers. \citet{khayatkhoeiEmergentAsymmetryPrecision} further addressed asymmetry issues through symPrecision and symRecall, while Precision Recall Cover introduced support-thresholded variants of precision and recall~\cite{cheemaPrecisionRecallCover}. Other alternatives include $\alpha$-Precision and $\beta$-Recall~\cite{alaaHowFaithfulYour} and TopP\&R~\cite{kimTopPRRobustSupport}, which further depart from the standard $k$NN construction. More recently, Clipped Density and Clipped Coverage were introduced as variants of Density and Coverage to improve interpretability by calibrating each sample's contribution~\cite{salvyEnhancedGenerativeModel2026}, while information-theoretic work has provided an entropy-based view of precision and recall, including additional quantities for evaluating mode collapse~\cite{foxUnifyingInformationtheoreticPerspective}.

\section{Discussion and Limitations}
\label{sec:discussion}

CADMA is evaluated primarily in image generation settings, where feature-space neighbourhoods are well established and widely used for generative model assessment. Although the framework is modality-agnostic at the level of finite sample sets, its behaviour in other domains such as text, graphs, or audio remains unexplored. Future work should investigate whether capacity-constrained matching remains informative under non-visual representations and domain-specific notions of similarity.

CADMA also relies on local neighbourhood construction in a pretrained feature space. Its diagnostic scores therefore depend on the quality of the feature extractor, the neighbourhood parameter $k$, and the balanced evaluation protocol used to compare real and generated samples. These choices are common in set-based generative evaluation, but they may affect the interpretation of mode dropping and mode collapse in practice.

From a broader-impact perspective, CADMA can improve transparency in generative model evaluation by revealing whether recall failures are driven by mode dropping or mode collapse. At the same time, it should not be treated as a complete replacement for human inspection or task-specific evaluation, since over-reliance on any single metric can still lead to misleading conclusions about model quality.

\section{Conclusion}
\label{sec:conclusion}

We studied recall-oriented evaluation for generative models and identified two limitations of existing geometry-based recall metrics: they can overestimate coverage under permissive neighbourhood relations, and they provide limited diagnostic information about coverage failures. To address these issues, we introduced CADMA, a capacity-aware framework based on local capacity matching. CADMA defines a capacity-aware coverage score $\mathcal{C}_{\mathrm{cap}}$ and two diagnostic scores, $\mathcal{D}_{\mathrm{drop}}$ and $\mathcal{D}_{\mathrm{collapse}}$, to quantify mode dropping and mode collapse. Across controlled Gaussian and CIFAR-10 experiments, CADMA follows the expected recall trends while providing diagnostic information that single-score metrics cannot reveal. The results suggest that capacity-constrained matching offers a useful way to move beyond permissive geometric coverage and toward more interpretable recall-style evaluation of generative models.

\section{Acknowledgements}

DM’s and GT’s work was funded by UK Research and Innovation (UKRI) under the UK government’s Horizon Europe funding guarantee (grant No. 10099264) and by the European Union (under EC Horizon Europe grant agreement No. 101135800 (RAIDO)). Yanran’s work was supported in part by the European Union under the Horizon Europe HORIZON-MSCA-2021-SE-01-01 project, Emerging Nanoscopy for Single Entity Characterisation (ENSIGN), Project No. 101086226. This research utilised Queen Mary’s Apocrita HPC facility, supported by QMUL Research-IT and ITS Research. For the purpose of open access, the author has applied a Creative Commons Attribution (CC BY) license to any Author Accepted Manuscript version arising.

{
    \small
    \bibliographystyle{references}
        \bibliography{references}
}

\appendix

\clearpage
\etocdepthtag.toc{supplementary}
\section*{Supplementary Material}

\noindent
Additional evaluations on ImageNet-256, FFHQ, LSUN Bedroom, and VAR
are presented in Appendix~\ref{app:models}.
Appendix~\ref{app:scalability} provides theoretical
time-complexity analysis and empirical runtime and memory measurements.

\begingroup
\etocsettagdepth{main}{none}
\etocsettagdepth{supplementary}{subsection}
\etocsettagdepth{after-supplementary}{none}
\etocsetnexttocdepth{subsection}
\etocsettocstyle{\subsection*{Contents}}{}
\tableofcontents
\endgroup

\clearpage

\section{Additional Method and Reproducibility Details}
\label{app:supplementary}

\subsection{Algorithmic Summary}
\label{app:algorithm}

This appendix summarizes the computation of CADMA. Given real features $\mathcal{R}=\{r_i\}_{i=1}^{N}$ and generated features $\mathcal{F}=\{f_j\}_{j=1}^{M}$, CADMA proceeds as follows:
\begin{enumerate}
    \item Compute the real-side $k$NN radius $\rho_i$ for each real feature $r_i$.
    \item Apply radius regularization by setting $\varepsilon_i=\min(\rho_i,\bar{\rho})$, where $\bar{\rho}=N^{-1}\sum_i \rho_i$.
    \item Construct the binary admissibility matrix $A$, where $A_{ij}=1$ if generated feature $f_j$ lies inside the regularized neighbourhood of real feature $r_i$.
    \item Solve the maximum-cardinality bipartite matching on $A$ to obtain $Z^\star$ and the matching size $N^{\mathrm{mat}}$.
    \item Report the capacity-aware coverage $\mathcal{C}_{\mathrm{cap}}=N^{\mathrm{mat}}/N$ and the diagnostic scores $\mathcal{D}_{\mathrm{drop}}$ and $\mathcal{D}_{\mathrm{collapse}}$ defined in Sec.~\ref{subsec:coverage_diagnostics}.
\end{enumerate}
The matching step is exact in our implementation. Different exact maximum-matching solvers may return different matching matrices when multiple optima exist, but they return the same matching size $N^{\mathrm{mat}}$. Consequently, for a fixed admissibility matrix $A$,
$\mathcal{C}_{\mathrm{cap}}$, $\mathcal{D}_{\mathrm{drop}}$, and
$\mathcal{D}_{\mathrm{collapse}}$ are well-defined and invariant
to the choice of maximum matching.

\subsection{Reproducibility Details}
\label{app:reproducibility}

All reported image-based experiments use DINOv2-Large features, as described in Sec.~\ref{sec:experiments}. For baseline metrics, we use the hyperparameters recommended by the corresponding papers whenever applicable. CADMA uses $k=70$ unless otherwise stated. The Gaussian experiments are generated synthetically from the distributions described in Sec.~\ref{sec:experiments}, while the CIFAR-10 experiments use public CIFAR-10 images with controlled class-level redistribution or outlier replacement.

The code release includes scripts for computing CADMA scores and reproducing the Gaussian experiments, together with the required dependencies and usage instructions.

\begin{figure*}[t]
    \centering
    \includegraphics[width=\linewidth]{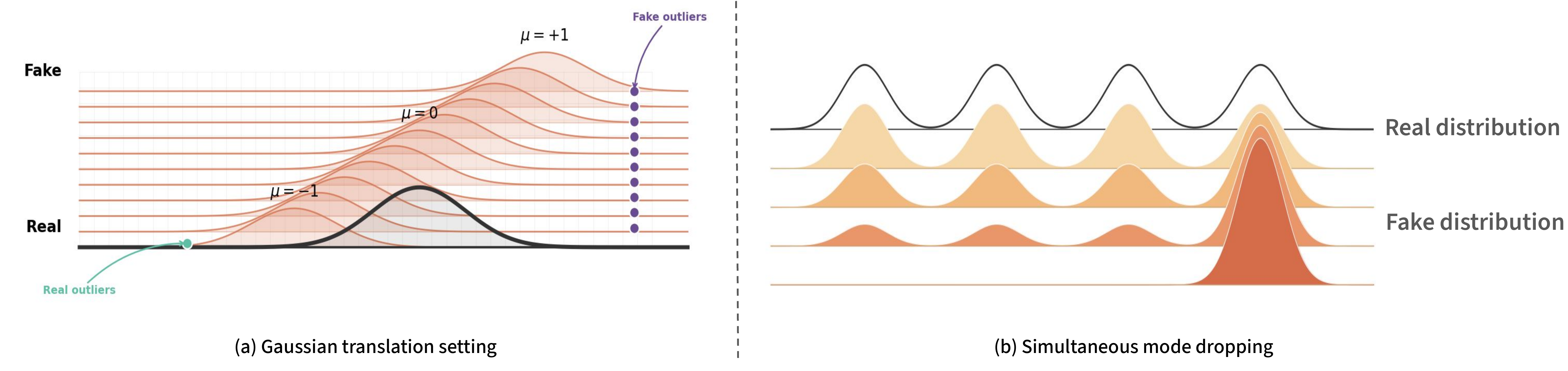}
    \caption{
    Supplementary illustration of the controlled experimental settings.
    Panel (a) shows the Gaussian translation setting with real and generated outliers, and panel (b) shows simultaneous mode dropping where generated mass is reassigned to the remaining mode.
    }
    \label{fig:supp_distribution_settings}
\end{figure*}

\subsection{Baseline Implementations}
\label{app:baseline_implementations}

We compare CADMA with a broad set of recall-style metrics. For implementation, we follow the original papers and public reference implementations whenever available:
\begin{itemize}
    \item Precision, Recall, Density, and Coverage~\cite{naeemReliableFidelityDiversity}: \url{https://github.com/clovaai/generative-evaluation-prdc}.
    \item symPrecision and symRecall~\cite{khayatkhoeiEmergentAsymmetryPrecision}: \url{https://github.com/mahyarkoy/emergent_asymmetry_pr}.
    \item P-precision and P-recall~\cite{parkProbabilisticPrecisionRecall2023}: \url{https://github.com/kdst-team/Probablistic_precision_recall}.
    \item TopP\&R~\cite{kimTopPRRobustSupport}: \url{https://github.com/LAIT-CVLab/TopPR}.
    \item Clipped Density and Clipped Coverage~\cite{salvyEnhancedGenerativeModel2026}: \url{https://github.com/nicolassalvy/ClippedDensityCoverage}.
\end{itemize}
For Precision Recall Cover~\cite{cheemaPrecisionRecallCover}, we follow the algorithmic description and recommended hyperparameters from the original paper. When a metric has method-specific hyperparameters, we use the recommended settings described in Sec.~\ref{sec:experiments}.

\subsection{Notes on Balanced Evaluation}
\label{app:balanced_evaluation}

The diagnostic scores are reported under balanced evaluation with $N=M$, following common practice in set-based generative evaluation. Using equally sized real and generated sets ensures that differences in sample counts do not distort the diagnostic scores. It is also natural for generative models, since one can usually sample the same number of generated examples as the real evaluation set. When the available sets have different sizes, a practical option is to evaluate CADMA over multiple balanced subsamples and average the resulting scores. This preserves the interpretation of $\mathcal{D}_{\mathrm{drop}}$ and $\mathcal{D}_{\mathrm{collapse}}$ as local imbalance diagnostics while avoiding direct comparison of unequal sample counts.

\section{Additional Evaluation on Existing Generative Models}
\label{app:models}

\subsection{Real-World Evaluation Setup}
\label{app:sample_sources}

We evaluate existing generative models on ImageNet-256, FFHQ, and LSUN Bedroom using publicly released generated samples from \citet{steinExposingFlawsGenerative}. These samples are also used in the evaluation of ClippedCoverage~\cite{salvyEnhancedGenerativeModel2026} and are available through the source paper's official code repository. For each model, we use the corresponding dataset-specific samples, with evaluation sizes fixed within each dataset: 25,000 for ImageNet-256, 50,000 for LSUN Bedroom, and 70,000 for FFHQ. For the large-scale generator evaluations on ImageNet-256, FFHQ, and LSUN Bedroom, we use $k=20$ for CADMA. For VAR~\cite{NEURIPS2024_9a24e284}, we instead generate samples using its official implementation, as detailed in Appendix~\ref{app:var}.

\subsection{ImageNet-256 Evaluation}
\label{app:imagenet}

To assess CADMA's diagnostic value on realistic large-scale data,
we evaluate 11 generators from diffusion, discrete-token, and GAN
families on class-conditional ImageNet-256. Each model is evaluated
using 25,000 real and 25,000 generated samples, with 25 samples
from each of the 1,000 categories in each set.

\begin{table}[tbp]
\centering
\caption{Evaluation of 11 generators on ImageNet-256. Each model uses 25,000 real and 25,000 generated samples.}
\label{tab:imagenet}
\small
\resizebox{\linewidth}{!}{\begin{tabular}{rllrrrr}
\toprule
\# & Family & Model (Year) & I-Precision $\uparrow$ & $\mathcal{C}_{\mathrm{cap}}$ $\uparrow$ & $D_{\mathrm{drop}}$ $\downarrow$ & $D_{\mathrm{collapse}}$ $\downarrow$ \\
\midrule
1 & Diffusion & ADM (2021) & 0.512 & 0.492 & 0.308 & 0.027 \\
2 & Diffusion & ADM-CFG (2021) & 0.589 & 0.674 & 0.164 & 0.034 \\
3 & Diffusion & ADM-CFG-ADMU (2021) & 0.619 & 0.710 & 0.146 & 0.039 \\
4 & Diffusion & LDM (2022) & 0.690 & 0.757 & 0.093 & 0.049 \\
5 & Diffusion & DiT-XL-2 (2023) & 0.731 & 0.786 & 0.087 & 0.042 \\
6 & Diffusion & DiT-XL-2-CFG (2023) & 0.926 & 0.826 & 0.035 & 0.145 \\
7 & Discrete-token & MaskGIT (2022) & 0.553 & 0.619 & 0.208 & 0.051 \\
8 & Discrete-token & RQ-Transformer (2022) & 0.453 & 0.410 & 0.323 & 0.032 \\
9 & GAN & BigGAN (2019) & 0.531 & 0.461 & 0.180 & 0.113 \\
10 & GAN & StyleGAN-XL (2022) & 0.504 & 0.571 & 0.220 & 0.057 \\
11 & GAN & GigaGAN (2023) & 0.525 & 0.549 & 0.244 & 0.044 \\
\bottomrule
\end{tabular}}
\end{table}
We make four observations. First, within the evaluated diffusion and GAN families, newer models generally achieve higher effective coverage. Specifically: $\mathcal{C}_{\mathrm{cap}}$ generally increases from earlier to newer diffusion models, and a similar improvement is observed from BigGAN to the more recent StyleGAN-XL and GigaGAN. GigaGAN is slightly below StyleGAN-XL, indicating that the trend is not strictly monotonic.

Second, among the evaluated models, diffusion models achieve the highest average $\mathcal{C}_{\mathrm{cap}}$ ($0.708$), compared with discrete-token generative models ($0.515$) and GANs ($0.527$). This result is consistent with the current dominance and strong empirical performance of diffusion-based approaches in image generation.

Third, GANs are widely known to be more susceptible to mode collapse, whereas diffusion models generally provide broader distribution coverage. Our results follow this expected pattern: the median $D_{\mathrm{collapse}}$ is $0.057$ for GANs and $0.041$ for diffusion models across all evaluated models. This agreement with the established behavior of GAN and diffusion models provides further evidence that $D_{\mathrm{collapse}}$ meaningfully captures mode concentration.

Finally, classifier-free guidance (CFG) is known to improve image fidelity and conditional alignment, but it can also concentrate generated samples around high-probability conditional modes. The improved image quality is first reflected in higher I-Precision. As more generated samples fall within the real-data domain, they increase the likelihood of covering additional valid regions of the real distribution and provide more candidates for matching. This can increase $\mathcal{C}_{\mathrm{cap}}$ and reduce $D_{\mathrm{drop}}$. Rows 5--6 show exactly this behavior: I-Precision increases from $0.731$ to $0.926$, $\mathcal{C}_{\mathrm{cap}}$ increases from $0.786$ to $0.826$, and $D_{\mathrm{drop}}$ decreases from $0.087$ to $0.035$. Meanwhile, the amplified conditioning signal concentrates more samples around conditional modes, increasing local density and redundancy; CADMA captures this effect through the increase of $D_{\mathrm{collapse}}$ from $0.042$ to $0.145$. Thus, the CADMA decomposition is consistent with both the expected improvement in fidelity and the expected increase in mode concentration caused by applying CFG.

\subsection{FFHQ Evaluation}
\label{app:ffhq}

On FFHQ, LDM achieves the best overall balance, with the highest $\mathcal{C}_{\mathrm{cap}}$ and lowest $D_{\mathrm{drop}}$. Efficient-VDVAE obtains the highest I-Precision but also the highest $D_{\mathrm{collapse}}$, indicating high fidelity but strong sample concentration. Conventional GANs generally exhibit higher mode collapse. Among them, StyleGAN-XL, which was designed to scale StyleGAN to diverse datasets, achieves the highest $\mathcal{C}_{\mathrm{cap}}$ and lowest $D_{\mathrm{collapse}}$. Transformer-based GANs perform better overall than conventional GAN baselines, with StyleNAT obtaining the second-highest $\mathcal{C}_{\mathrm{cap}}$ and the lowest $D_{\mathrm{collapse}}$.
\begin{table}[tbp]
\centering
\caption{Evaluation of nine generators on FFHQ.
Best and second-best results are bolded and underlined, respectively.}
\label{tab:ffhq}
\small
\resizebox{\linewidth}{!}{\begin{tabular}{rllrrrr}
\toprule
\# & Family & Model (Year) & I-Precision $\uparrow$ & $\mathcal{C}_{\mathrm{cap}}$ $\uparrow$ & $D_{\mathrm{drop}}$ $\downarrow$ & $D_{\mathrm{collapse}}$ $\downarrow$ \\
\midrule
1 & VAE & Efficient-VDVAE (2022) & \textbf{0.448} & 0.417 & $\underline{0.154}$ & 0.223 \\
2 & Diffusion & LDM (2022) & $\underline{0.430}$ & \textbf{0.606} & \textbf{0.143} & $\underline{0.067}$ \\
3 & Discrete-token & Unleashing Transformers (2022) & 0.285 & 0.395 & 0.209 & 0.102 \\
4 & GAN & StyleGAN2-ADA (2020) & 0.159 & 0.283 & 0.190 & 0.129 \\
5 & GAN & InsGen (2021) & 0.203 & 0.355 & 0.218 & 0.101 \\
6 & GAN & Projected-GAN (2021) & 0.109 & 0.191 & 0.170 & 0.107 \\
7 & GAN & StyleGAN-XL (2022) & 0.328 & 0.481 & 0.204 & 0.068 \\
8 & Transformer-GAN & StyleSwin (2022) & 0.328 & 0.486 & 0.202 & 0.085 \\
9 & Transformer-GAN & StyleNAT (2022) & 0.389 & $\underline{0.561}$ & 0.185 & \textbf{0.054} \\
\bottomrule
\end{tabular}}
\end{table}

\subsection{LSUN Bedroom Evaluation}
\label{app:lsun}

On LSUN Bedroom, diffusion models again achieve the strongest results. ADMNet-dropout ranks first on all four metrics, while iDDPM obtains the second-highest $\mathcal{C}_{\mathrm{cap}}$ and second-lowest $D_{\mathrm{collapse}}$. The Consistency Model achieves competitive I-Precision but substantially worse distributional coverage, with high $D_{\mathrm{drop}}$ and $D_{\mathrm{collapse}}$. Since consistency models target fast few-step generation, this result suggests that sampling efficiency may come with reduced coverage and stronger sample concentration in this setting. GAN-based and diffusion--GAN models also generally exhibit substantially higher $D_{\mathrm{collapse}}$ than standard diffusion models.
\begin{table}[tbp]
\centering
\caption{Evaluation of eight generators on LSUN Bedroom.
Best and second-best results are bolded and underlined, respectively.}
\label{tab:lsun}
\small
\resizebox{\linewidth}{!}{\begin{tabular}{rllrrrr}
\toprule
\# & Family & Model (Year) & I-Precision $\uparrow$ & $\mathcal{C}_{\mathrm{cap}}$ $\uparrow$ & $D_{\mathrm{drop}}$ $\downarrow$ & $D_{\mathrm{collapse}}$ $\downarrow$ \\
\midrule
1 & Diffusion & DDPM (2020) & 0.306 & 0.439 & 0.260 & 0.031 \\
2 & Diffusion & iDDPM (2021) & 0.383 & $\underline{0.530}$ & 0.201 & $\underline{0.028}$ \\
3 & Diffusion & ADMNet-dropout (2021) & \textbf{0.521} & \textbf{0.687} & \textbf{0.085} & \textbf{0.015} \\
4 & Consistency & Consistency (2023) & 0.377 & 0.288 & 0.282 & 0.268 \\
5 & Discrete-token & Unleashing Transformers (2022) & 0.209 & 0.234 & 0.305 & 0.108 \\
6 & GAN & StyleGAN (2019) & $\underline{0.418}$ & 0.465 & 0.217 & 0.111 \\
7 & GAN & Projected-GAN (2021) & 0.222 & 0.109 & $\underline{0.179}$ & 0.240 \\
8 & Diffusion--GAN hybrid & Diffusion-Projected-GAN & 0.224 & 0.148 & 0.215 & 0.215 \\
\bottomrule
\end{tabular}}
\end{table}

\subsection{VAR Evaluation and Sampling Strategies}
\label{app:var}

We evaluate VAR at four different depths---d16, d20, d24, and d30---using 25,000 class-balanced ImageNet-256 samples, with 25 samples per category. We additionally compare VAR with DiT-XL-2 using the standard CFG scale of $1.5$ from the original DiT paper. For clarity, the DiT-XL-2-CFG model reported in our previous ImageNet experiment is a high-CFG variant using scale $4.0$, rather than the standard CFG setting used here.

\begin{table}[tbp]
\centering
\caption{VAR evaluation on ImageNet-256 under original and smooth
sampling, with DiT-XL-2 at CFG scale 1.5 as a reference.
Bold values indicate the better result between the two sampling
strategies at each VAR depth.}
\label{tab:var}
\small
\begin{tabular}{llrrrr}
\toprule
Model (Year) & Sampling & I-Precision $\uparrow$ & $\mathcal{C}_{\mathrm{cap}}$ $\uparrow$ & $D_{\mathrm{drop}}$ $\downarrow$ & $D_{\mathrm{collapse}}$ $\downarrow$ \\
\midrule
VAR-d16 (2024) & Original & 0.636 & 0.703 & 0.139 & \textbf{0.039} \\
VAR-d16 (2024) & Smooth & \textbf{0.758} & \textbf{0.798} & \textbf{0.059} & 0.069 \\
VAR-d20 (2024) & Original & 0.694 & 0.769 & 0.103 & \textbf{0.040} \\
VAR-d20 (2024) & Smooth & \textbf{0.819} & \textbf{0.832} & \textbf{0.043} & 0.067 \\
VAR-d24 (2024) & Original & 0.735 & 0.800 & 0.085 & \textbf{0.036} \\
VAR-d24 (2024) & Smooth & \textbf{0.844} & \textbf{0.849} & \textbf{0.038} & 0.068 \\
VAR-d30 (2024) & Original & 0.726 & 0.795 & 0.095 & \textbf{0.034} \\
VAR-d30 (2024) & Smooth & \textbf{0.824} & \textbf{0.846} & \textbf{0.048} & 0.058 \\
DiT-XL-2 (2023) & CFG=1.5 & 0.731 & 0.786 & 0.087 & 0.042 \\
\bottomrule
\end{tabular}
\end{table}

Under original sampling, increasing VAR depth from d16 to d24 consistently improves I-Precision and $\mathcal{C}_{\mathrm{cap}}$ and reduces $\mathcal{D}_{\mathrm{drop}}$. This improvement does
                           not continue from d24 to d30: $\mathcal{C}_{\mathrm{cap}}$ slightly decreases from 0.800 to 0.795 and $\mathcal{D}_{\mathrm{drop}}$ increases from 0.085 to 0.095, although $\mathcal{D}_{\mathrm{collapse}}$ decreases.
                           Compared with DiT-XL-2, VAR-d24 provides only a marginal improvement in I-Precision (0.735 vs. 0.731), while its advantage is more evident in the
                           diversity-related diagnostics: $\mathcal{C}_{\mathrm{cap}}$ increases from 0.786 to 0.800, and $\mathcal{D}_{\mathrm{collapse}}$ decreases from 0.042 to 0.036. Thus, CADMA shows that VAR's
                           improvement over DiT lies more in effective coverage and mode balance than in sample fidelity.
                           We additionally evaluate VAR's smooth sampling option. It consistently improves I-Precision and $\mathcal{C}_{\mathrm{cap}}$ and reduces $\mathcal{D}_{\mathrm{drop}}$, showing that more generated
                           samples fall within the real-data domain and contribute to valid matching. However, it also increases $\mathcal{D}_{\mathrm{collapse}}$, revealing stronger local sample
                           concentration.

\subsection{Cross-Model Observations}
\label{app:observations}

These results support three observations. First, diffusion models generally dominate the evaluated methods and provide strong effective coverage: LDM achieves the highest $\mathcal{C}_{\mathrm{cap}}$ on FFHQ, while ADMNet-dropout and iDDPM lead on LSUN Bedroom. Second, high fidelity allows more generated samples to fall within the real-data domain, but these samples may concentrate around a limited number of modes rather than provide balanced coverage. Efficient-VDVAE achieves the highest FFHQ I-Precision but has substantially lower $\mathcal{C}_{\mathrm{cap}}$ and higher $D_{\mathrm{collapse}}$ than LDM. Third, some efficiency-oriented generators appear to sacrifice diversity: Consistency exhibits high $D_{\mathrm{collapse}}$, while Unleashing Transformers exhibits high $D_{\mathrm{drop}}$.

\section{Computational Cost and Scalability}
\label{app:scalability}

\subsection{Theoretical Time Complexity}
\label{app:complexity}

Let both real and generated sets contain $n$ samples with
$d$-dimensional features. Under a brute-force exact-distance implementation, exact neighborhood construction in the high-dimensional setting considered here requires $T_{\mathrm{Coverage}}=O(dn^2)$ time ~\cite{liu2004investigation}.
CADMA additionally constructs a sparse bipartite graph with $E$ admissible
edges in $O(n+E)$ time and computes maximum-cardinality matching in
$O(E\sqrt{n})$ time using the classical Hopcroft--Karp bound~\cite{hopcroft1973n}. Computing $D_{\mathrm{drop}}$ and
$D_{\mathrm{collapse}}$ requires only $O(n+E)$ additional time. Hence,

\[
T_{\mathrm{CADMA}}=O(dn^2+n+E+E\sqrt{n})=O(dn^2+E\sqrt{n}).
\]

In our experiments, the admissibility graphs $A$ induced by local
$k$-NN neighborhoods remain sparse, with $k=70\ll n$ for
$1{,}000\leq n\leq 50{,}000$. Under the sparse-graph assumption
$E=O(kn)$, the time complexity reduces to

\[
T_{\mathrm{CADMA}}=O(dn^2+kn^{3/2}).
\]

In this situation, the time cost is dominated by neighborhood construction,
$O(dn^2)$, similarly to $T_{\mathrm{Coverage}}$.

As the number of samples increases, the relative additional time cost $\rho_{\mathrm{HK}}^{\mathrm{time}}$ is

\[
\rho_{\mathrm{HK}}^{\mathrm{time}}=\frac{T_{\mathrm{matching}}}{T_{\mathrm{neighborhood}}}=O(\frac{kn^{3/2}}{dn^2})=O(\frac{k}{d\sqrt{n}}).
\]

For fixed $k$ and $d$, this ratio approaches zero as $n$ grows
under the sparse-graph assumption $E=O(kn)$. Therefore, the additional matching cost is asymptotically negligible relative to neighborhood construction, consistent with the matching-runtime analysis in the main paper, which reports that Hopcroft--Karp matching accounts for only approximately $1\%$ of the total runtime.

\subsection{Wall-Clock Runtime}
\label{app:runtime}

Feature-space neighborhood construction is involved not only in CADMA but also in most neighborhood-based recall-style metrics. The CADMA column reports the complete runtime, $T_{\mathrm{neigh}}$ reports feature-neighborhood construction time, and $T_{\mathrm{add}}$ includes Hopcroft--Karp matching and score decomposition. Across 10K--50K samples, the operations included in $T_{\mathrm{add}}$ account for only $0.262\%$--$1.189\%$ of the total runtime. Over the same 10K--50K range, CADMA requires $74.6\%$--$81.2\%$ of the runtime of ClippedCoverage and $98.3\%$--$107.7\%$ of that of classical Coverage. Thus, CADMA is faster than ClippedCoverage and comparable to Coverage, despite using $k=70$ rather than $k=5$ for Coverage.

\begin{table}[tbp]
\centering
\caption{Wall-clock runtime of CADMA and baseline metrics, measured
in seconds using precomputed features. $N$ denotes the number of
samples in each of the real and generated sets.}
\label{tab:runtime}
\small
\resizebox{\linewidth}{!}{\begin{tabular}{rllrrrr}
\toprule
$N$ & CADMA (s) & $T_{\mathrm{neigh}}$ (s) & $T_{\mathrm{add}}$ (s) & $T_{\mathrm{add}}/T_{\mathrm{CADMA}}$ & ClippedCoverage (s) & Coverage (s) \\
\midrule
1K & 0.225 & 0.223427 & 0.001189 & 0.529\% & 0.205 & 0.158 \\
2.5K & 0.662 & 0.658441 & 0.003792 & 0.572\% & 0.795 & 0.625 \\
5K & 2.166 & 2.157869 & 0.007753 & 0.358\% & 2.641 & 2.094 \\
7.5K & 4.792 & 4.779245 & 0.012223 & 0.255\% & 5.674 & 4.466 \\
10K & 8.266 & 8.230676 & 0.034747 & 0.420\% & 10.252 & 7.947 \\
20K & 31.757 & 31.645776 & 0.110717 & 0.349\% & 39.798 & 30.817 \\
30K & 68.509 & 68.328158 & 0.179587 & 0.262\% & 84.795 & 69.686 \\
40K & 123.505 & 122.035564 & 1.468471 & 1.189\% & 165.461 & 116.686 \\
50K & 193.375 & 191.806797 & 1.566866 & 0.810\% & 238.140 & 179.521 \\
\bottomrule
\end{tabular}}
\end{table}
The runtimes were measured on a node with two 24-core Intel Xeon Platinum 8268 CPUs (2.90 GHz) and an NVIDIA A100 PCIe 40 GB GPU, using one CPU thread and precomputed features.

\subsection{Memory Usage}
\label{app:memory}

As shown below, feature storage is the dominant memory cost, requiring $970.1$ MB for 50K samples, whereas CADMA increases peak memory by only $6.8\%$, making the additional memory overhead practically minor compared with feature storage.

Here, Feature Memory denotes the memory used after loading the features but before metric computation. The remaining columns report the additional peak memory introduced by each metric, with percentages relative to Feature Memory. All memory values are reported in MB.

\begin{table}[tbp]
\centering
\caption{Memory usage in MB. Percentages in parentheses are
relative to the Feature Memory column.}
\label{tab:memory}
\small
\begin{tabular}{lrrrr}
\toprule
$N$ & Feature Memory & CADMA & ClippedCoverage & Coverage \\
\midrule
1K & 167.3 & 8.3 (5.0\%) & 3.5 (2.1\%) & 5.2 (3.1\%) \\
2.5K & 191.6 & 11.1 (5.8\%) & 3.7 (1.9\%) & 9.9 (5.2\%) \\
5K & 234.0 & 16.8 (7.3\%) & 6.1 (2.6\%) & 16.3 (6.9\%) \\
7.5K & 274.7 & 24.4 (8.9\%) & 4.1 (1.5\%) & 17.0 (6.2\%) \\
10K & 314.3 & 28.5 (9.1\%) & 3.9 (1.3\%) & 15.7 (5.0\%) \\
20K & 478.7 & 39.5 (8.3\%) & 10.2 (2.1\%) & 24.2 (5.1\%) \\
30K & 640.6 & 50.0 (7.8\%) & 12.9 (2.0\%) & 32.5 (5.1\%) \\
40K & 806.0 & 52.8 (6.6\%) & 19.2 (2.4\%) & 40.4 (5.0\%) \\
50K & 970.1 & 65.8 (6.8\%) & 21.2 (2.2\%) & 53.0 (5.4\%) \\
\bottomrule
\end{tabular}\end{table}

\etocdepthtag.toc{after-supplementary}

\newpage
\section*{NeurIPS Paper Checklist}

\begin{enumerate}

\item {\bf Claims}
    \item[] Question: Do the main claims made in the abstract and introduction accurately reflect the paper's contributions and scope?
    \item[] Answer: \answerYes{}     \item[] Justification: The main claims are stated in the abstract and introduction and are supported by the method in Sec.~\ref{sec:method} and experiments in Sec.~\ref{sec:experiments}. We scope the claims to recall-oriented evaluation and diagnosis for generative models.


\item {\bf Limitations}
    \item[] Question: Does the paper discuss the limitations of the work performed by the authors?
    \item[] Answer: \answerYes{}     \item[] Justification: The paper discusses limitations in Sec.~\ref{sec:discussion}, including feature-space dependence, and evaluation beyond image generation.

\item {\bf Theory assumptions and proofs}
    \item[] Question: For each theoretical result, does the paper provide the full set of assumptions and a complete (and correct) proof?
    \item[] Answer: \answerNA{}     \item[] Justification: The paper does not present formal theoretical results requiring proofs. CADMA is defined algorithmically in Sec.~\ref{sec:method}.


    \item {\bf Experimental result reproducibility}
    \item[] Question: Does the paper fully disclose all the information needed to reproduce the main experimental results of the paper to the extent that it affects the main claims and/or conclusions of the paper (regardless of whether the code and data are provided or not)?
    \item[] Answer: \answerYes{}     \item[] Justification: The method, baselines, datasets, feature representations, and evaluation protocols are described in Secs.~\ref{sec:method} and~\ref{sec:experiments}. Additional implementation details are provided in the code.


\item {\bf Open access to data and code}
    \item[] Question: Does the paper provide open access to the data and code, with sufficient instructions to faithfully reproduce the main experimental results, as described in supplemental material?
    \item[] Answer: \answerYes{}     \item[] Justification: We provide code with documentation, dependencies, and scripts for computing CADMA and reproducing experiments.

\item {\bf Experimental setting/details}
    \item[] Question: Does the paper specify all the training and test details (e.g., data splits, hyperparameters, how they were chosen, type of optimizer) necessary to understand the results?
    \item[] Answer: \answerYes{}     \item[] Justification: We specify the datasets, generated sample sets, feature extractor, baselines, and CADMA hyperparameters in Sec.~\ref{sec:experiments}.


\item {\bf Experiment statistical significance}
    \item[] Question: Does the paper report error bars suitably and correctly defined or other appropriate information about the statistical significance of the experiments?
    \item[] Answer: \answerNo{}     \item[] Justification: We do not report error bars for all experiments. Instead, we use controlled comparisons and multiple settings to support the main claims.


\item {\bf Experiments compute resources}
    \item[] Question: For each experiment, does the paper provide sufficient information on the computer resources (type of compute workers, memory, time of execution) needed to reproduce the experiments?
    \item[] Answer: \answerYes{}     \item[] Justification: The main text reports the hardware used for the experiments and includes runtime analysis for CADMA matching in Sec.~\ref{sec:experiments}.


\item {\bf Code of ethics}
    \item[] Question: Does the research conducted in the paper conform, in every respect, with the NeurIPS Code of Ethics \url{https://neurips.cc/public/EthicsGuidelines}?
    \item[] Answer: \answerYes{}     \item[] Justification: The authors have reviewed the NeurIPS Code of Ethics, and the work uses public datasets and generated samples without human-subject experiments or private data.


\item {\bf Broader impacts}
    \item[] Question: Does the paper discuss both potential positive societal impacts and negative societal impacts of the work performed?
    \item[] Answer: \answerYes{}     \item[] Justification: We discuss potential positive and negative impacts in Sec.~\ref{sec:discussion}, including improved transparency in generative model evaluation and the risk of over-reliance on any single metric.


\item {\bf Safeguards}
    \item[] Question: Does the paper describe safeguards that have been put in place for responsible release of data or models that have a high risk for misuse (e.g., pre-trained language models, image generators, or scraped datasets)?
    \item[] Answer: \answerNA{}     \item[] Justification: The paper does not release high-risk datasets or generative models. CADMA is an evaluation method and does not itself generate content.


\item {\bf Licenses for existing assets}
    \item[] Question: Are the creators or original owners of assets (e.g., code, data, models), used in the paper, properly credited and are the license and terms of use explicitly mentioned and properly respected?
    \item[] Answer: \answerYes{}     \item[] Justification: We cite the datasets, generated sample sources, pretrained feature extractors, and baseline methods used in the experiments, and follow their stated licenses and terms of use.


\item {\bf New assets}
    \item[] Question: Are new assets introduced in the paper well documented and is the documentation provided alongside the assets?
    \item[] Answer: \answerYes{}     \item[] Justification: We release CADMA code as a new asset with documentation, dependencies, usage instructions, and licensing information.


\item {\bf Crowdsourcing and research with human subjects}
    \item[] Question: For crowdsourcing experiments and research with human subjects, does the paper include the full text of instructions given to participants and screenshots, if applicable, as well as details about compensation (if any)?
    \item[] Answer: \answerNA{}     \item[] Justification: The paper does not involve crowdsourcing experiments or research with human subjects.


\item {\bf Institutional review board (IRB) approvals or equivalent for research with human subjects}
    \item[] Question: Does the paper describe potential risks incurred by study participants, whether such risks were disclosed to the subjects, and whether Institutional Review Board (IRB) approvals (or an equivalent approval/review based on the requirements of your country or institution) were obtained?
    \item[] Answer: \answerNA{}     \item[] Justification: The paper does not involve crowdsourcing experiments or research with human subjects, so IRB approval or equivalent review is not applicable.


\item {\bf Declaration of LLM usage}
    \item[] Question: Does the paper describe the usage of LLMs if it is an important, original, or non-standard component of the core methods in this research? Note that if the LLM is used only for writing, editing, or formatting purposes and does \emph{not} impact the core methodology, scientific rigor, or originality of the research, declaration is not required.
        \item[] Answer: \answerNA{}     \item[] Justification: The core method development and experiments do not involve LLMs as important, original, or non-standard components.


\end{enumerate}

\end{document}